\pdfoutput=1
\documentclass{article}
\usepackage{spconf,amsmath,graphicx,multirow,booktabs}

\title{Adversarial Examples for Good: Adversarial Examples Guided Imbalanced Learning}
\name{Jie Zhang$^{\star}$ \qquad Lei Zhang$^{\star}$ \qquad Gang Li \qquad Chao Wu$^{\dagger}$ \thanks{$^{\star}$ Jie Zhang and Lei Zhang contributed equally.} \thanks{ $^{ \dagger}$  Corresponding author, chao.wu@zju.edu.cn.}}

\address{Zhejiang University, China}
\begin{document}
%
\maketitle
\begin{abstract}
Adversarial examples are inputs for machine learning models that have been designed by attackers to cause the model to make mistakes.
In this paper, we demonstrate that adversarial examples can also be utilized for good to improve the performance of imbalanced learning. We provide a new perspective on how to deal with imbalanced data: adjust the biased decision boundary by training with Guiding Adversarial Examples (GAEs). Our method can effectively increase the accuracy of minority classes while sacrificing little accuracy on majority classes. We empirically show, on several benchmark datasets, our proposed method is comparable to the state-of-the-art method. To our best knowledge,  we are the first to deal with imbalanced learning with adversarial examples.

\end{abstract}
\section{Introduction}
\label{sec:introduction}

In practical, most of real-world datasets have long-tailed label distributions~\cite{liu2019large}.  As a result, deep learning algorithms usually perform poorly or even collapse when the training data suffers from heavy class-imbalance, especially for highly skewed data~\cite{zhang2021bag}. 
Due to the imbalanced data, networks can be over-fitting to the minority classes, which leads to the deviation of the learned classification boundary~\cite{cao2019learning}. 

Recent research aims to modify the data sampler to balance class frequency during optimization (re-sampling), and modify the weights of the classification loss to increase the importance of tail classes (re-weighting). In general, re-weighting and re-sampling are two main approaches that are commonly used to learn long-tailed data~\cite{cao2019learning}.  In contrast to previous works, we provide a new perspective on how to deal with imbalanced data:  adjust the biased decision boundary by training with Guiding Adversarial Examples (GAEs).\par

\begin{figure}[t]
    \centering
    \centerline{\includegraphics[width=8cm]{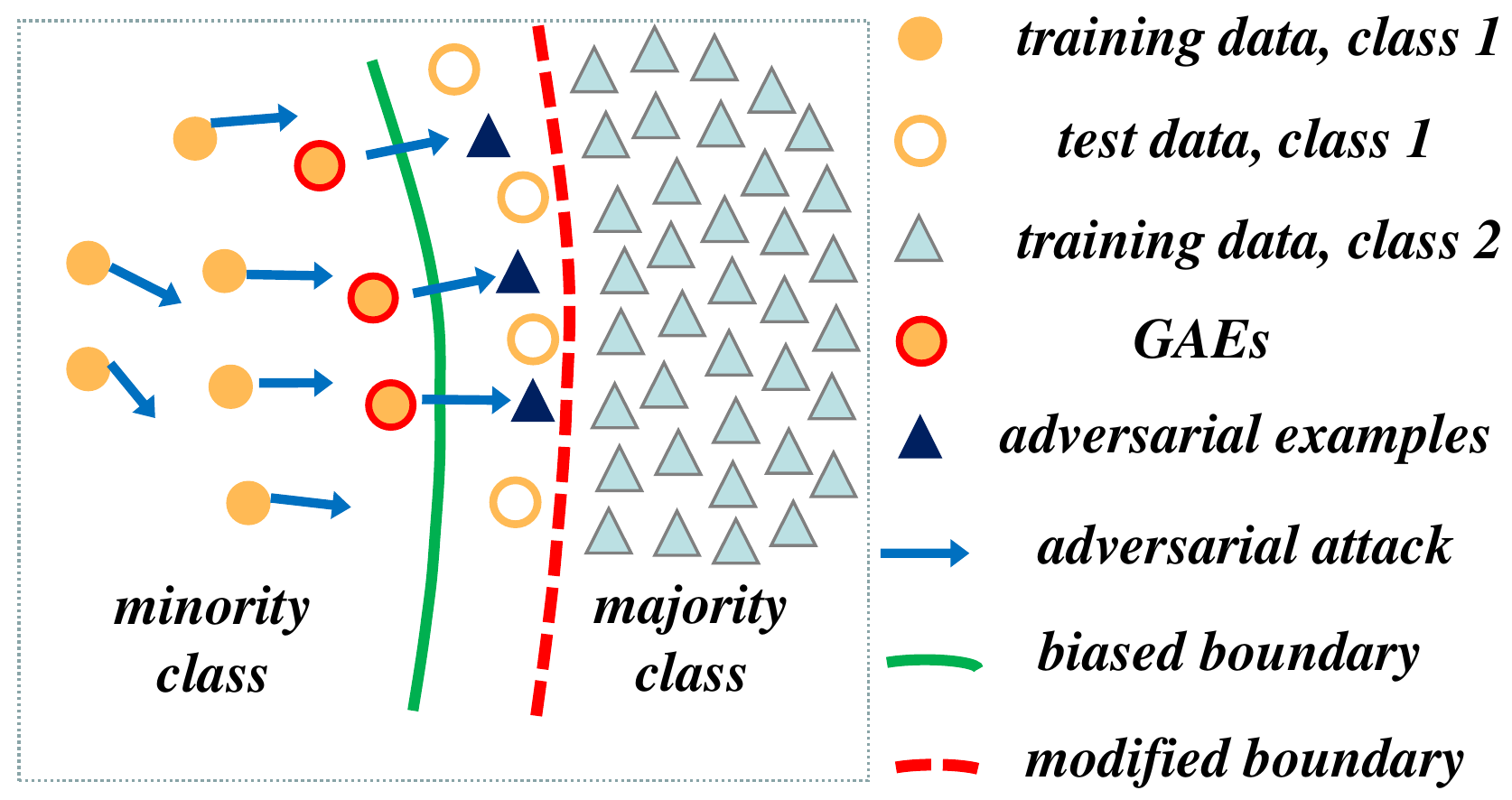}}
  \caption{An illustration of Guiding Adversarial Examples.}
  \label{fig1}
  \vspace{-0.3cm}
\end{figure}
\begin{figure}[t]
    \centering
    \centerline{\includegraphics[width=7.5cm]{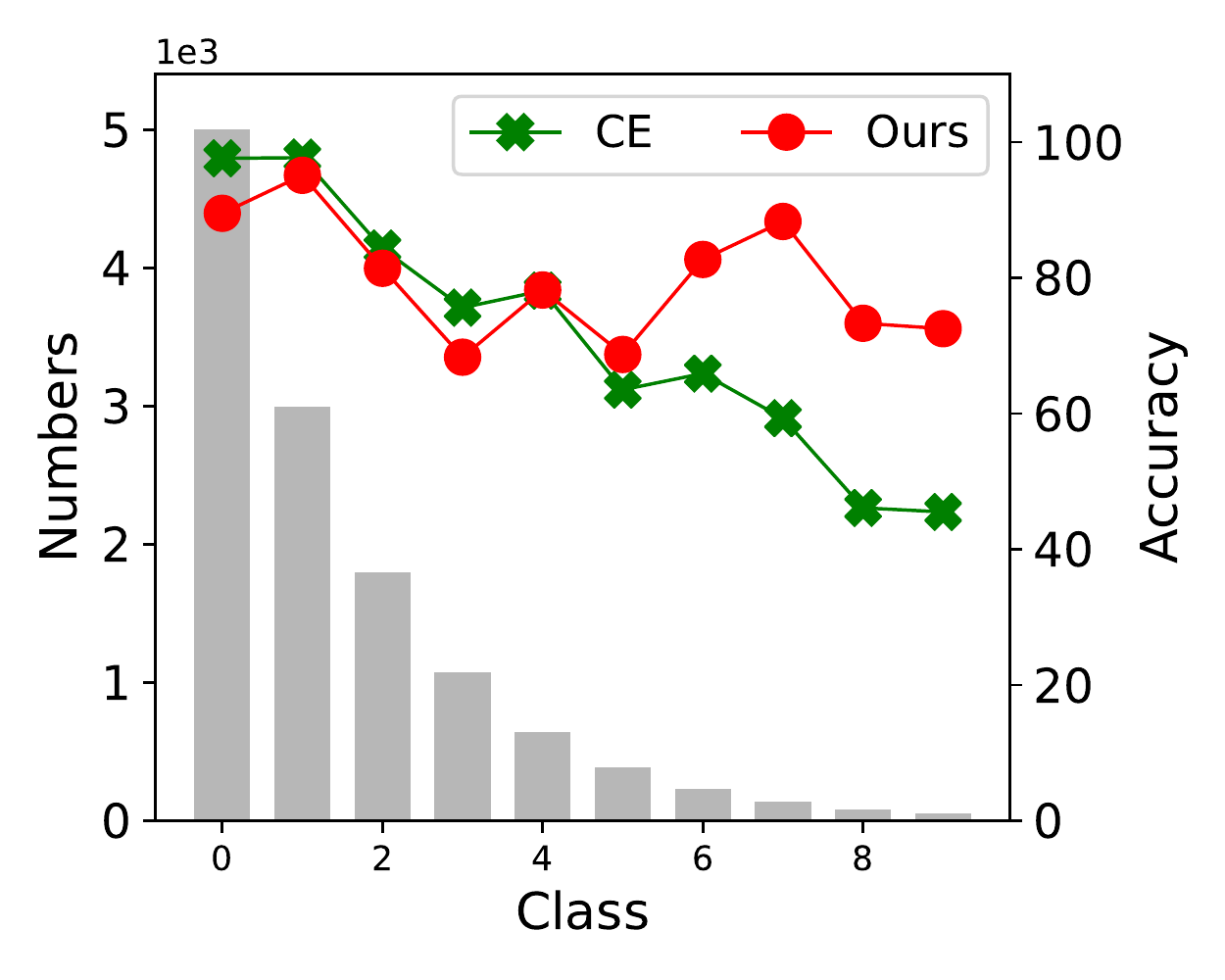}}
    \vspace{-0.6cm}
  \caption{Top-1 accuracy for each class on CIFAR10.}
  
  \label{fig2}
  \vspace{-0.5cm}
\end{figure}

Adversarial examples are inputs to neural networks that are deliberately designed to cause the networks to misclassify~\cite{goodfellow2014explaining,huang2022cmua,liu2022efficient,zhang2022towards}. Prior studies of adversarial examples have primarily focused on robustness, which indicates the vulnerability of models against adversarial attacks~\cite{Salman2019ProvablyRD,zhang2019theoretically,chen2022towards}. Rather than focusing on the robustness of models, we present an interesting use of adversarial examples to increase the performance of models on long-tailed data. Interestingly, in addition to attacking and reducing the accuracy of a model, adversarial samples can also be employed to enhance its accuracy. To begin with, we show a  definition of Guiding Adversarial Examples in Figure~\ref{fig1}, which refers to adversarial examples in minority classes that can be transferred to majority classes within a few steps. These examples can further be utilized to guide the learning of the biased decision boundary.

To get examples beneficial to guiding the learning, we modify an attack scheme to search examples closed to the biased decision boundary.  For the boundary-oriented adjustment, we generate GAEs that have just crossed the decision boundary with a few steps to guide the learning of the model. As illustrated in Figure~\ref{fig2}, the accuracy on plain models (train with Cross-Entropy) drops from head to tail, which is exactly what traditional long-tailed recognition aims to solve. It is worth noting that our method can effectively increase the accuracy of minority classes while sacrificing little accuracy on majority classes.
\par

Our main contributions are summarized as follows:\par

\begin{itemize}
\item[$\bullet$] To our best knowledge,  we are the first to deal with imbalanced learning with adversarial examples. We demonstrate that adversarial examples can also be utilized for good to improve the performance of imbalanced learning.
\item[$\bullet$] Numerous experiments demonstrate the superiority of
our approach which is as effective as the state-of-the-art algorithms on various datasets.
\end{itemize}

\section{Related Works}
\label{sec:rw}


\subsection{Learning from Imbalanced Data}
\label{ssec:lfid}

There have been many studies focusing on analyzing imbalanced data in recent years~\cite{he2009learning,liu2019large,zhang2021bag}. 
It is usually the case that real-world data exhibits an imbalanced distribution, and highly skewed data can adversely affect the effectiveness of machine learning~\cite{jamal2020rethinking,cao2019learning}. 
Re-sampling~\cite{chawla2002smote} and re-weighting~\cite{kim2020adjusting,jamal2020rethinking,cui2019class} 
are traditional methods towards addressing imbalanced data. 
Through re-weighting strategies, modern works can make networks more sensitive to minority categories by assigning a variable weight to each class. 
Moreover, two re-sampling methods, under-sampling frequent classes and over-sampling minority classes, have been extensively discussed in previous studies. Also, we saw success with new perspectives, such as deferred re-balancing 
~\cite{cao2019learning} schedule and decoupled training ~\cite{kang2019decoupling}. 


\subsection{Adversarial Examples}
\label{ssec:ae}

Christian Szegedy et al \cite{szegedy2013intriguing} 
refer to the image which is added with small noise that cannot be perceived by human 
but makes the output of neural network change explicitly 
as "adversarial examples". 
For object recognition, the predicted label of the adversarial example is different 
from that of the original example. 
Adversarial training technique had been introduced in 
\cite{goodfellow2014explaining}, 
and adversarial examples are commonly used in
optimizing the adversarial robustness of neural networks. 
It is worth noting that adversarial examples are originally utilized 
to attack neural networks. 
Aleksander Madry et al ~\cite{madry2017towards} use multi-step projected gradient descent 
(PGD) as an attack method to generate adversarial examples to train the model and 
the robustness of the result model is improved substantially. 
However, adversarial examples techniques have rarely benefited other areas. In this paper, We demonstrate that adversarial examples can also be utilized for good to improve the performance of imbalanced learning. 

\begin{figure*}[htb]
  \centering
  \includegraphics[width=16.0cm]{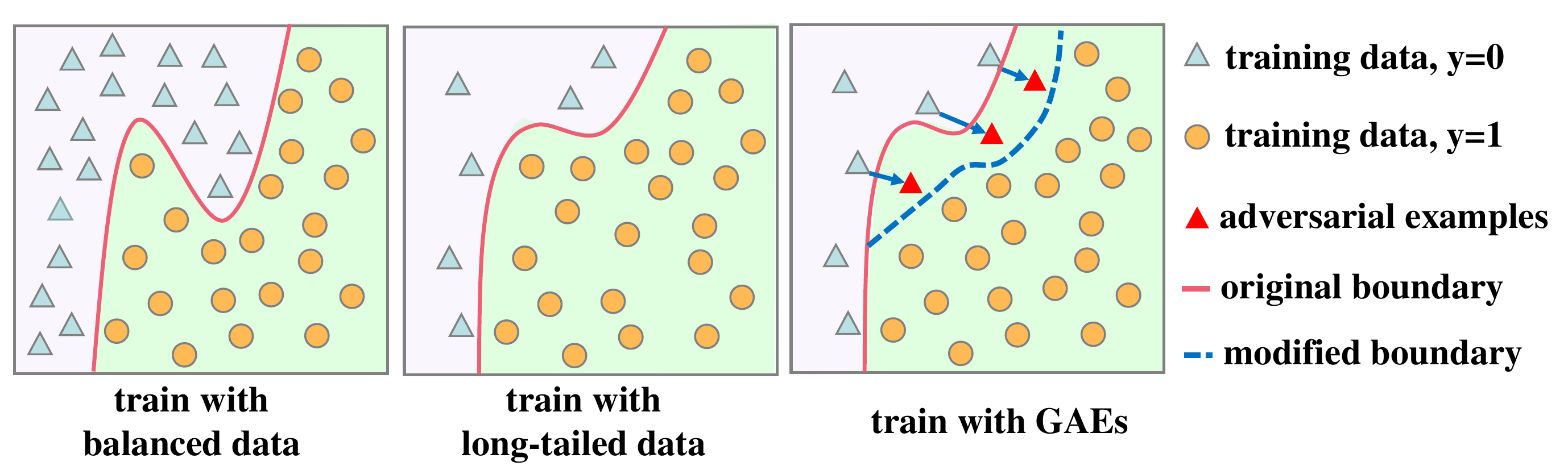}
  \caption{An illustration of our method. By training with GAEs, the biased decision boundary can be futher corrected.}
  \label{fig:the_priciple_of_method}
  \vspace{-0.5cm}
\end{figure*}

\section{Method}
\label{sec:method}

In this section, we start with the concept of guiding adversarial examples (GAEs), 
and then describe the proposed method to benefit long-tailed classification. 

\subsection{Guiding Adversarial Examples}
\label{ssec:GAEs}

The projected gradient descent (PGD) \cite{madry2017towards} we adopted is the most common method for finding adversarial examples. Given $x^{0} \in \mathcal{X}$,
\begin{equation}
  x^{t+1}=\Pi_{\mathcal{B}[x^0]}\left(x^t+\alpha \, \text{sgn} (\nabla_{x^t} L(\theta,x^t,y))\right)
\end{equation}
until a threshold is met. 
$x^0$ denotes an original example and $\Pi_{\mathcal{B}[x^0]}\left( \cdot \right)$ is the projection function. 
Since PGD is a multi-step method, the hyper-parameter $k$ denotes the number of steps. 
we refer to adversarial examples in minority classes that can be transferred to majority classes within a few steps as the Guiding Adversarial Examples (GAEs). These examples can guide the learning of the biased decision boundary while other adversarial examples have little guiding significance. 
It is important to note that with too many steps, the biased decision boundary will be {over-corrected}. 
Therefore, we limit the number of steps $k$ to a small integer (usually less than or equal to 3). 

The GAEs are generated from the tail examples which are close to the biased decision boundary that separates minority classes from majority classes. 
And we only calculate the adversarial examples for the tail examples $(x_{tail},y_{tail})$ with $k$ steps. 
Let $(x,y)$ denote a mini-batch of examples in minority classes, and $(x_{adv},y_{adv})$ is the corresponding  adversarial examples.
The confidence scores for the true label and misclassified label are represented by $f_t(x)$ and $f_m(x)$, respectively. 
In this way, we use PGD to generate adversarial examples as:
\begin{equation} \begin{aligned}
s=i+1,\,\text{where} \; i+1\leq k \; \text{and} \\
\begin{cases}
\, f_t(x_{adv}^i)-f_m(x_{adv}^i)>0 \\
\, f_t(x_{adv}^{i+1})-f_m(x_{adv}^{i+1})<0
\end{cases}
,
\end{aligned} \end{equation}
where $s$ means the number of steps required for a successful attack. 
However, the adversarial examples may be located in other minority classes, which will restrict the effect of improving accuracy. 
Consequently, not all adversarial examples among $(x_{adv},y_{adv})$ participate in the training, but the adversarial examples $(x_{cross},y_{cross})$ with frequent classes after the attacks. As the classification boundary becomes more balanced during training, the number of GAEs should gradually decrease.

\subsection{Adversarial Examples Guided Learning}
\label{ssec:AEGL}
GAEs are utilized to direct the learning of a convergent and more or less unbalanced model, from which we generate GAEs, trained in some way on an unbalanced dataset. 
To simplify the procedure, Our criterion is merely the cross-entropy loss
\begin{equation}
L_{CE}(q,p)=-\sum_i^mp(x_i) \log q(x_i).
\end{equation}
We believe combining effective training techniques with our method will result in better performance, and we will futher investigate it in the future. 
In the following is the cross-entropy loss aiming to guide the training of the model's decision boundary:

\begin{equation}
L_{cross}=L_{CE}(h(x_{cross}),y_{tail}^{'})
\end{equation}
where $h(x)$ are the softmax output of a model on the input $x$ and $y_{tail}^{'}$ denote the original labels of $x_{cross}$.


\section{Experiments}
\label{sec:experiments}

\begin{table*}[!ht]
  \centering
    \caption{Accuracy (\%) on imbalanced SVHN, FashionMNIST and CIFAR100.}
  \begin{tabular}{c|cccc|cccc|cccc}
  \toprule
    Datasets & \multicolumn{4}{c|}{SVHN} & \multicolumn{4}{c|}{FashionMNIST} & \multicolumn{4}{c}{CIFAR100} \\ \midrule
    Imbalance ratio $\rho$ & {0.1} & {0.05} & {0.02} & {0.01} &  {0.1} & {0.05} & {0.02} & {0.01} & {0.1} & {0.05} & {0.02} & {0.01} \\ \midrule
    CE & 93.93&92.21&90.12&87.41 & 88.89&88.25&86.37&85.35 & 56.29&51.24&43.27&38.17 \\ \midrule
    \textbf{Ours}&\textbf{94.40}&\textbf{93.18}&\textbf{91.78}&\textbf{89.57} & \textbf{89.52}&\textbf{88.95}&\textbf{87.18}&\textbf{85.85} & \textbf{58.22}&\textbf{53.19}&\textbf{45.80}&\textbf{39.93} \\ \bottomrule
  \end{tabular}

  \label{tb2}
\end{table*}

\begin{figure}[htb]
\begin{minipage}[b]{.48\linewidth}
  \centering
  \centerline{\includegraphics[width=4.0cm]{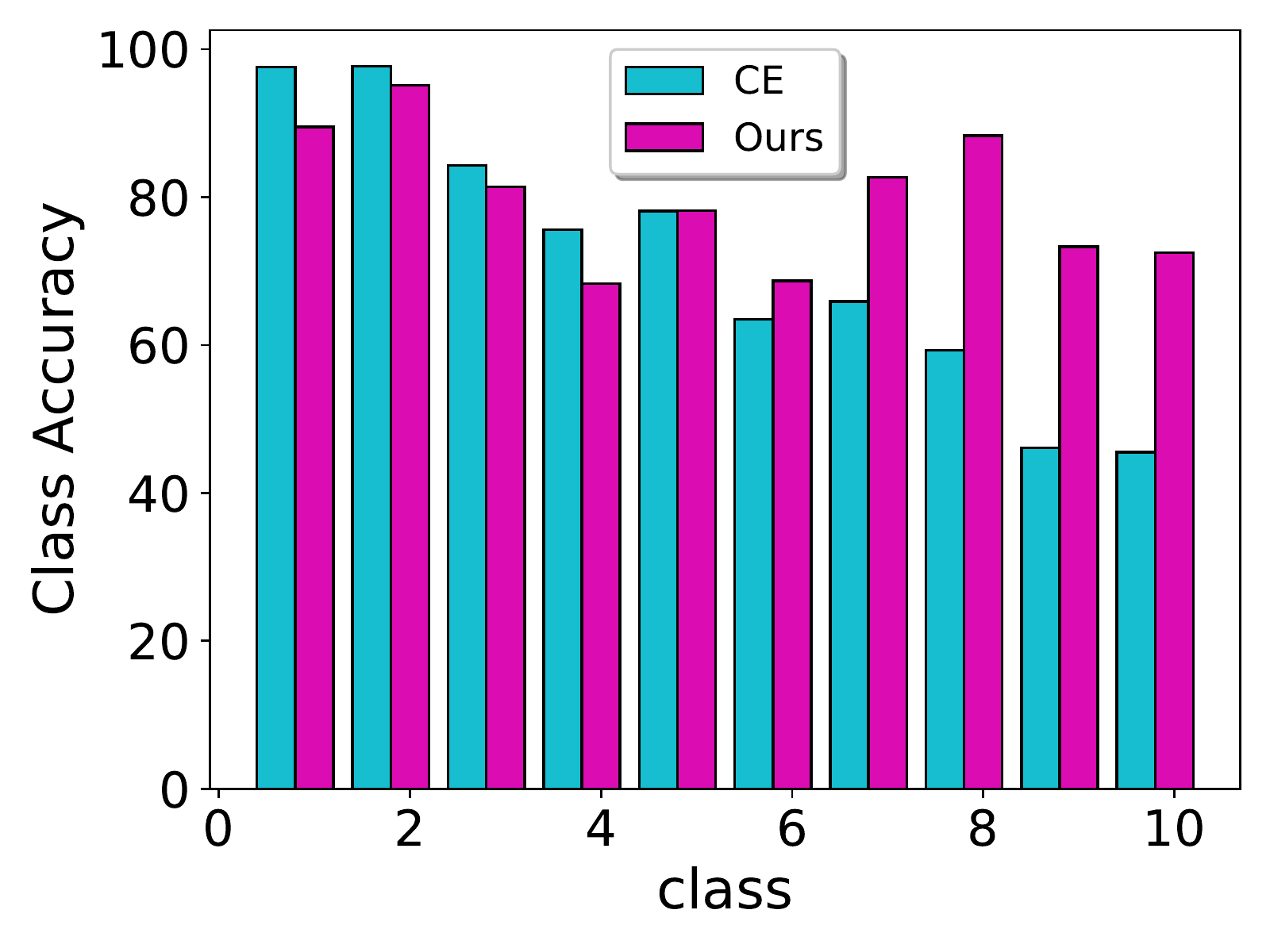}}
  \centerline{(a) On CIFAR10.}\medskip
\end{minipage}
\hfill
\begin{minipage}[b]{0.48\linewidth}
  \centering
  \centerline{\includegraphics[width=4.0cm]{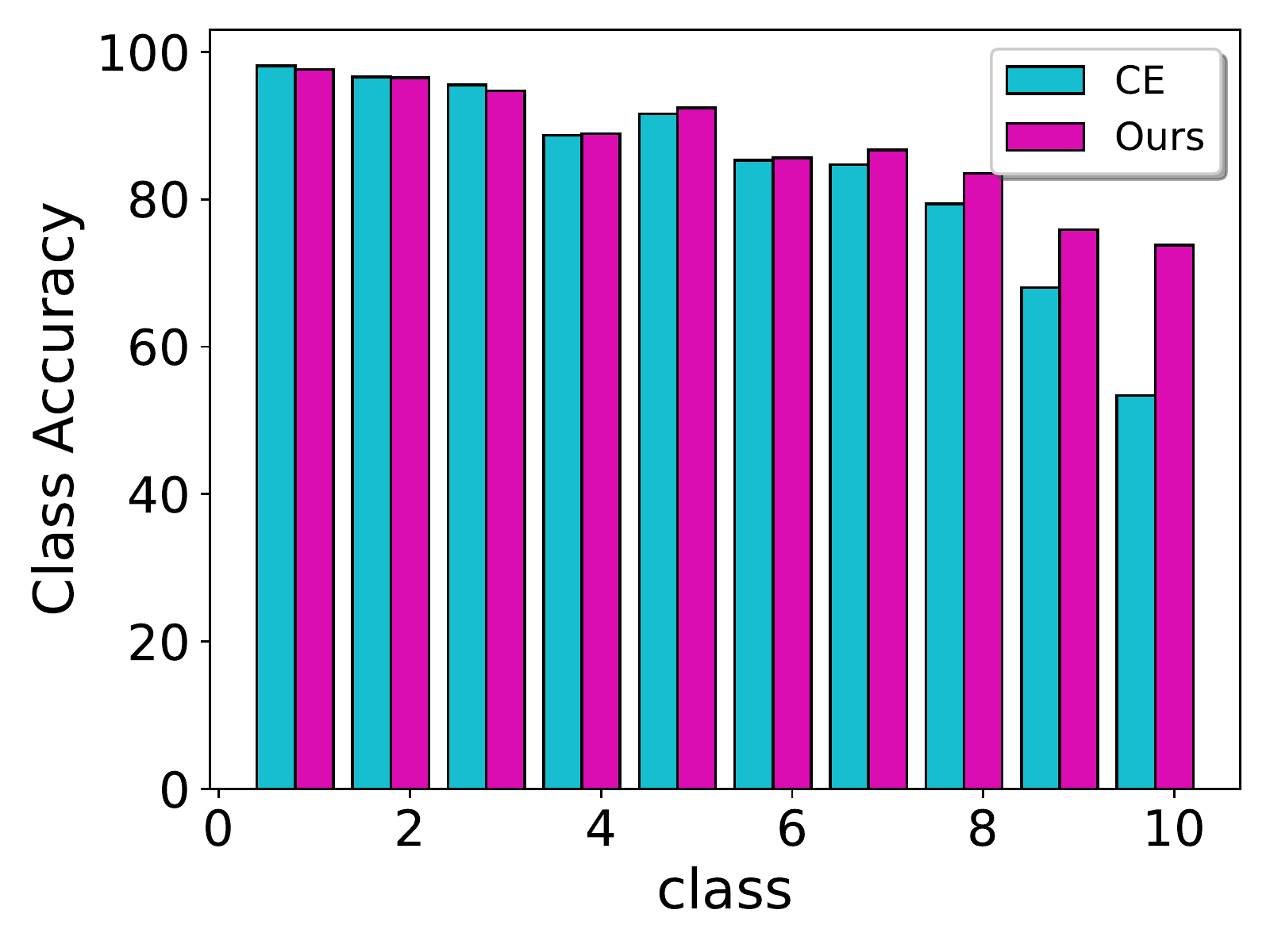}}
  \centerline{(b) On SVHN.}\medskip
\end{minipage}
\caption{Comparison of accuracy on each class.}
\label{fig_exp1}
\vspace{-0.5cm}
\end{figure}

\begin{figure}[htb]
  \begin{minipage}[b]{1.0\linewidth}
    \centering
    \centerline{\includegraphics[width=8.5cm]{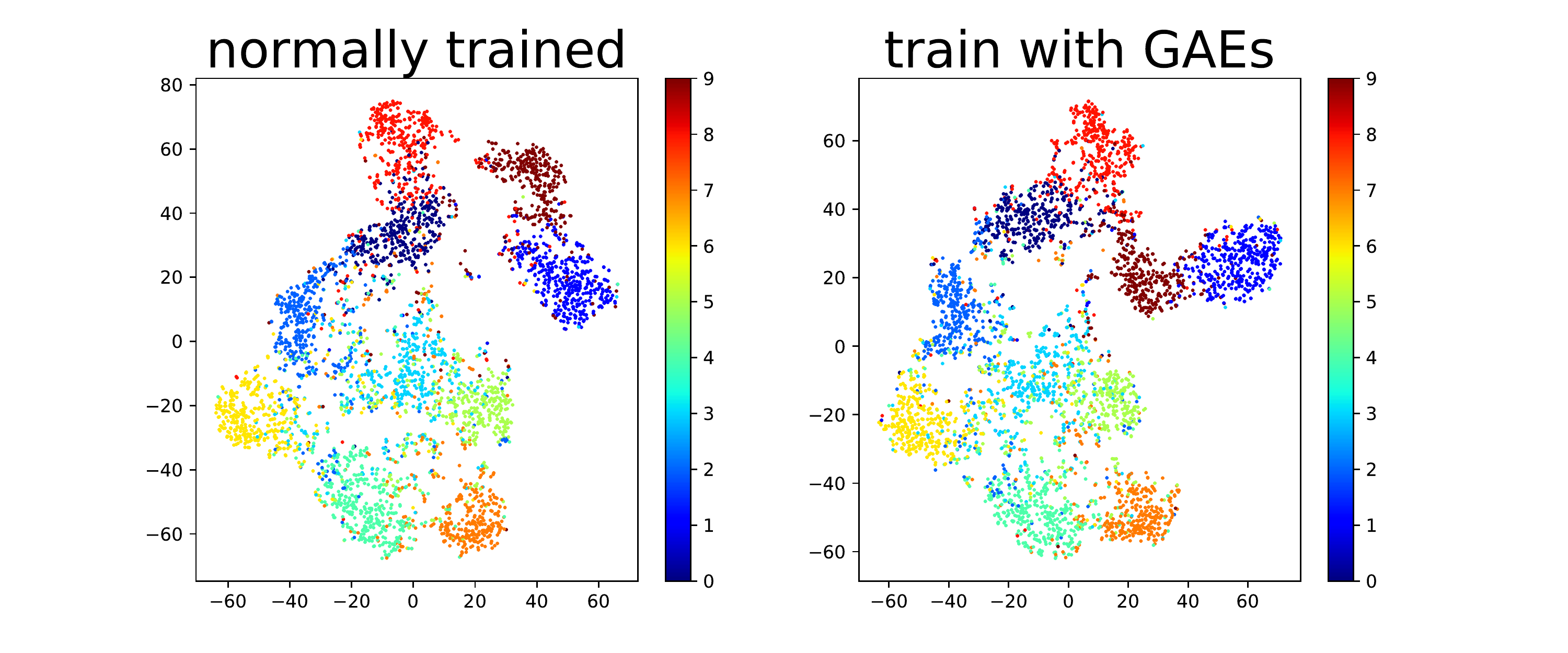}}
  \end{minipage}
  \caption{T-sne visulization of standard trainning and our method On CIFAR10. The test accuracy is 71.36\% and 80.23\%, respectively. }
  \label{tsne}
\vspace{-0.5cm}
\end{figure}
\begin{figure}[htb]
  \begin{minipage}[b]{1.0\linewidth}
    \centering
    \centerline{\includegraphics[width=7.5cm]{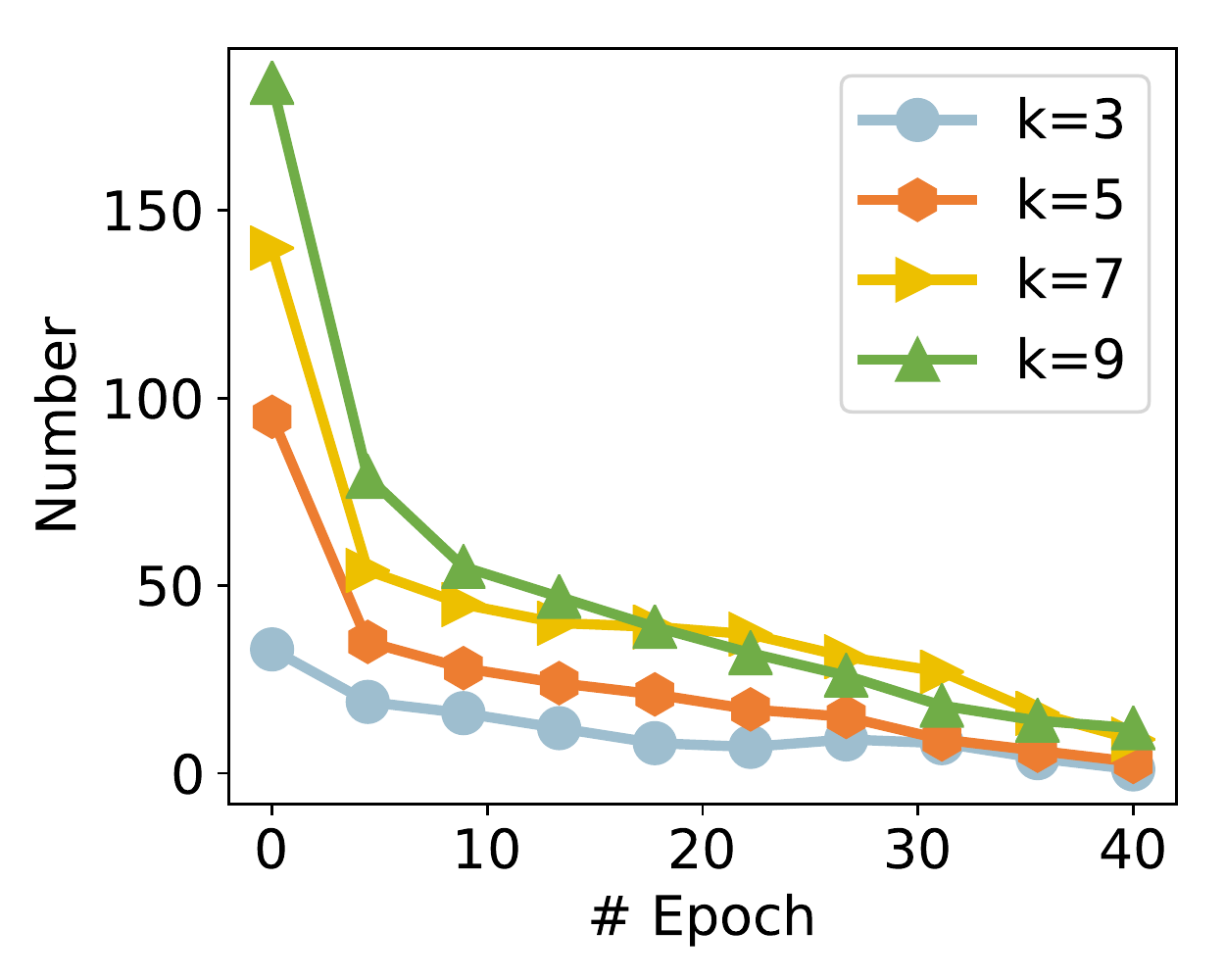}}
  \end{minipage}
  \vspace{-1cm}
  \caption{The number of GAEs during training.}
  \label{fig5}
  \vspace{-0.5cm}
\end{figure}
\subsection{Experiment Setups}

In this study, we conducted a number of experiments on four popular image classification benchmark datasets: 
SVHN~\cite{netzer2011reading}, FashionMNIST~\cite{xiao2017fashion}, CIFAR-10 and CIFAR-100~\cite{krizhevsky2009learning} with controllable degrees of data imbalance. Follow the setting in~\cite{cao2019learning},  the magnitude of the long-tailed imbalance declines exponentially with class size. The imbalance ratio $\rho$ represents the ratio between the samples of the least and most frequent classes. In this paper, we compare our method to standard training (using the Cross-Entropy (CE) loss function) as well as the state-of-the-art LDAM-DRW~\cite{cao2019learning}.
Despite the fact  that state-of-the-art two-stage approaches~\cite{kang2019decoupling} have achieved high accuracy; nevertheless, the baselines we use is sufficient for our needs, as our goal is to
evaluate the effectiveness of our method, not achieve the best possible accuracy on each task. 

\begin{table}
  \centering
    \caption{Accuracy (\%) on imbalanced CIFAR-10.}
  \begin{tabular}{c|c|c|c}
  \toprule
    Imbalance ratio $\rho$ & {0.01} & {0.02} & {0.1} \\ 
    \midrule
    CE&71.36&76.89&86.76 \\ \midrule
    CE+DRW&76.01&80.11&87.35 \\ \midrule
    LDAM&73.72&78.98&86.34 \\ \midrule
    LDAM+DRW&77.83&81.67&87.74 \\ \midrule
    \textbf{Ours}&\textbf{80.23}&\textbf{81.85}&\textbf{87.86} \\ 
    \bottomrule
  \end{tabular}

  \label{tb1}
  \vspace{-0.3cm}
\end{table}

\begin{table}
\centering
\caption{Ablation study for the effect of $k$.}
\label{tb3}
\begin{tabular}{cccccc}
\toprule
Steps & 1 & 3 & 5 & 7 & 9 \\
\midrule
Ours($\rho$=0.01) & 74.11 & \textbf{80.23} & 77.81 & 76.74 & 70.15 \\
\midrule
Ours($\rho$=0.02) & 76.98 & \textbf{81.85} & 79.48 & 79.01 & 75.41 \\
\midrule
Ours($\rho$=0.1) & 86.57 & \textbf{87.86} & 86.48 & 86.16 & 85.14 \\
\bottomrule
\end{tabular}
\vspace{-0.3cm}
\end{table}
\subsection{Experimental Results}
\label{ssec:er}
We use a CNN architecture as our initial model on FashionMNIST and  use ResNet-32 on other datasets. A popular re-weighting strategy, DRW~\cite{cao2019learning}, is also used in our comparisons. We first report the results on CIFAR10 dataset in Table~\ref{tb1}. We demonstrate that adversarial samples can also be utilized for good to improve the performance of imbalanced learning by a large margin, e.g.\ the accuracy increases from 71.36\% to 80.23\% when the imbalance ratio $\rho=0.01$. We also demonstrate that our algorithm consistently outperforms the previous state-of-the-art methods.

More detailed evaluations on other datasets can be seen in Table~\ref{tb2}. We report the top-1 accuracy on four datasets under various imbalanced settings. As shown in Table~\ref{tb2}, our method demonstrates significant improvements over the standard training, even on larger datasets CIFAR100 with 100 classes. Similarly, our method achieves significant improvements in accuracy by 2\% for the complex dataset CIFAR100. It is also worth mentioning that when the training data becomes increasingly unbalanced, our method is becoming increasingly effective. We would like to argue that this is reasonable. As the more imbalanced data can result in a more biased decision boundary, which is exactly what our method attempts to tackle.

Furthermore, we present the test accuracy on each classes of SVHN and CIFAR10 in Figure~\ref{fig_exp1}. Obviously, we find that our method can significantly increase the accuracy of minority classes while only sacrificing little accuracy on majority classes. A T-sne visualization of the standard training and our method can be seen in Figure~\ref{tsne}, which indicates that training with GAEs can produce a more accurate decision boundary.

\subsection{Ablation Study}
Our ablation study begins by investigating the contribution of the number of steps introduced in our method to generate GAEs. We vary the steps $k=\{1,3,5,7,9\}$ and report the corresponding top-1 accuracy on imbalanced CIFAR10 dataset in Table~\ref{tb3}. Clearly, the number of steps required to generate GAEs directly impacts the results of model training. A small $k=1$ will  slightly correct the classification boundary with a few GAEs. When a large value of k is used (e.g.\ k=9), our method can generate too many adversarial examples, resulting in an over-correction of the decision boundary, which reduces the performance of the model. We also report the number of GAEs during training for different number of steps as an additional support for our method. In Figure~\ref{fig5}, we demonstrate that a larger $k$ can lead to more GAEs and during training the number of GAEs will decrease. The explanation for this phenomenon is that the decision boundary is constantly being adjusted, and thus the number of effective GAEs is declining.

\section{Conclusion}
\label{sec:conclusion}
In this work,  we demonstrate that adversarial examples can also be utilized for good to improve the performance of imbalanced learning. To our best knowledge,  we are the first to deal with imbalanced learning with adversarial examples. On several benchmark datasets, our proposed method is comparable to the state-of-the-art method.

\section{Acknowledgement}
This work was supported by the National Key Research and Development Project of China (2021ZD0110400 No. 2018AAA0101900), National Natural Science Foundation of China (U19B2042), The University Synergy Innovation Program of Anhui Province (GXXT-2021-004), Zhejiang Lab (2021KE0AC02), Academy Of Social Governance Zhejiang University, Fundamental Research Funds for the Central Universities (226-2022-00064), Artificial Intelligence Research Foundation of Baidu Inc., Program of ZJU and Tongdun Joint Research Lab.

\vfill\pagebreak


\bibliographystyle{IEEEbib}
\bibliography{refs.bib}

\end{document}